\pdfoutput=1
\PassOptionsToPackage{table}{xcolor}  % must be before acl.sty loads

\documentclass[11pt]{article}

 \usepackage[preprint]{acl}

\usepackage{times}
\usepackage{amsmath}
\usepackage{latexsym}
\usepackage{todonotes}
\usepackage{changepage}
\usepackage[most]{tcolorbox}       % nice coloured boxes
\usepackage{fontawesome5}          % icons (trophy, bolt, chess-knight)
\usepackage{enumitem} % For proper rendering and hyphenation of words containing Latin characters (including in bib files)
\usepackage[T1]{fontenc}
\usepackage[utf8]{inputenc}
\usepackage{tcolorbox}
\usepackage{epigraph}
\usepackage{microtype}

\usepackage{inconsolata}

\usepackage{graphicx}

\usepackage{algorithm}
\usepackage{algorithmic}

\usepackage{cleveref}   % \Cref for figures, tables, algs, eqs

\usepackage{booktabs}   % professional-looking tables
\usepackage{multirow}   % table cells spanning rows
\usepackage{array}      % extended tabular column definitions

\usepackage{float}

\title{Do We Need Large VLMs for Spotting Soccer Actions?}

\author{%
  \begin{tabular}{cc}
    Ritabrata Chakraborty\thanks{Corresponding author} &
    Rajatsubhra Chakraborty \\[0.3em]
    Manipal University Jaipur                   &
    UNC–Charlotte                               \\[1.2em]
    Avijit Dasgupta        & Sandeep Chaurasia \\[0.3em]
    IIIT Hyderabad         & Manipal University Jaipur \\[2.0em]
    \multicolumn{2}{c}{\texttt{ritabrata.229301716@muj.manipal.edu}}
  \end{tabular}
}

\begin{document}
\maketitle
\begin{abstract}
Traditional video-based tasks like soccer action spotting rely heavily on visual inputs, often requiring complex and computationally expensive models to process dense video data. We propose a shift from this video-centric approach to a text-based task, making it lightweight and scalable by utilizing Large Language Models (LLMs) instead of Vision-Language Models (VLMs). We posit that expert commentary, which provides rich descriptions and contextual cues contains sufficient information to reliably spot key actions in a match. To demonstrate this, we employ a system of three LLMs acting as judges specializing in outcome, excitement, and tactics for spotting actions in soccer matches. Our experiments show that this language-centric approach performs effectively in detecting critical match events coming close to state-of-the-art video-based spotters while using zero video processing compute and similar amount of time to process the entire match. 
\end{abstract}

\section{Introduction}
\epigraph{Football is a game of mistakes. Whoever makes the fewest mistakes wins.}{Johan Cruyff}

In the domain of video understanding \cite{nguyen-etal-2024-video}, visual frames have traditionally been considered the best input for many tasks, including action spotting, event detection, and object recognition \cite{Giancola_2025,Giancola_2023_CVPR,8553793} . However, these methods often require significant computational resources to process and analyze the dense video data \cite{Selva_2023,feichtenhofer2019slowfast}. Despite the advancements in video models, such as convolutional neural networks (CNNs) \cite{karpathy2014large} and vision transformers (ViTs), the need for high-resolution video inputs can be prohibitive in both training and deployment scenarios.
% In many video contexts, particularly in sports, there is a spoken commentary that accompanies the visual content. This commentary provides detailed descriptions of the ongoing actions, events, and emotions, offering an alternative source of information. It is intriguing to consider whether these spoken descriptions, which can often provide rich contextual cues, could replace or complement visual inputs for specific tasks. By focusing on the textual information provided by commentators, we may gain a lightweight, scalable solution that avoids the complexity of video-based models.

Action spotting \cite{seweryn2023survey}, a core task in sports analytics, aims to identify key events within a video, such as goals, penalties, or substitutions, by analyzing the visual content. Manual methods by broadcasters were slow and took time in distribution \cite{merler}.Traditional approaches \cite{shih2017survey} have relied on object detection and tracking techniques that require parsing every frame of the video to detect specific actions \cite{khan2018soccer}. These methods can be computationally expensive and often struggle with long sequences or multiple simultaneous events \cite{xu2025actionspottingpreciseevent}. In contrast, when considering the commentary, each moment in the match is often described in rich detail, including the action, the players involved, and the contextual relevance. The spoken word can provide a nuanced understanding of the match dynamics, capturing moments of excitement, controversy, and strategic importance that may not always be fully conveyed through visual data alone. This raises an interesting possibility: \textit{\textbf{\textcolor{blue}{Can we leverage textual commentary as a primary input for action spotting, bypassing the need for video frames?}}}

We explore this question by proposing a text based action spotting pipeline using an LLM-as-a-judge setup, following \cite{zheng2023judging}. We investigate whether expert commentary is enough for current LLMs to infer actions from, and if it is comparable to heavy video based action spotter VLMs. We also study the improvement in action spotting as time taken per match and the independence from video processing compute.To this end, we provide the following contributions:
% In this paper, we explore this question by converting the traditional video-based action spotting task into a text-based one. Specifically, we investigate whether expert commentary, which provides fine-grained descriptions and contextual cues, is sufficient for reliably spotting key actions. The use of Large Language Models (LLMs) enables us to process and evaluate these textual descriptions, which contain detailed insights into match events, without requiring visual data. This approach not only alleviates the need for computationally intensive video models but also opens up new avenues for action detection in contexts where visual data may be scarce or of low quality, such as archived footage or radio broadcasts.
\begin{figure}
    \centering
    \includegraphics[width=1\linewidth]{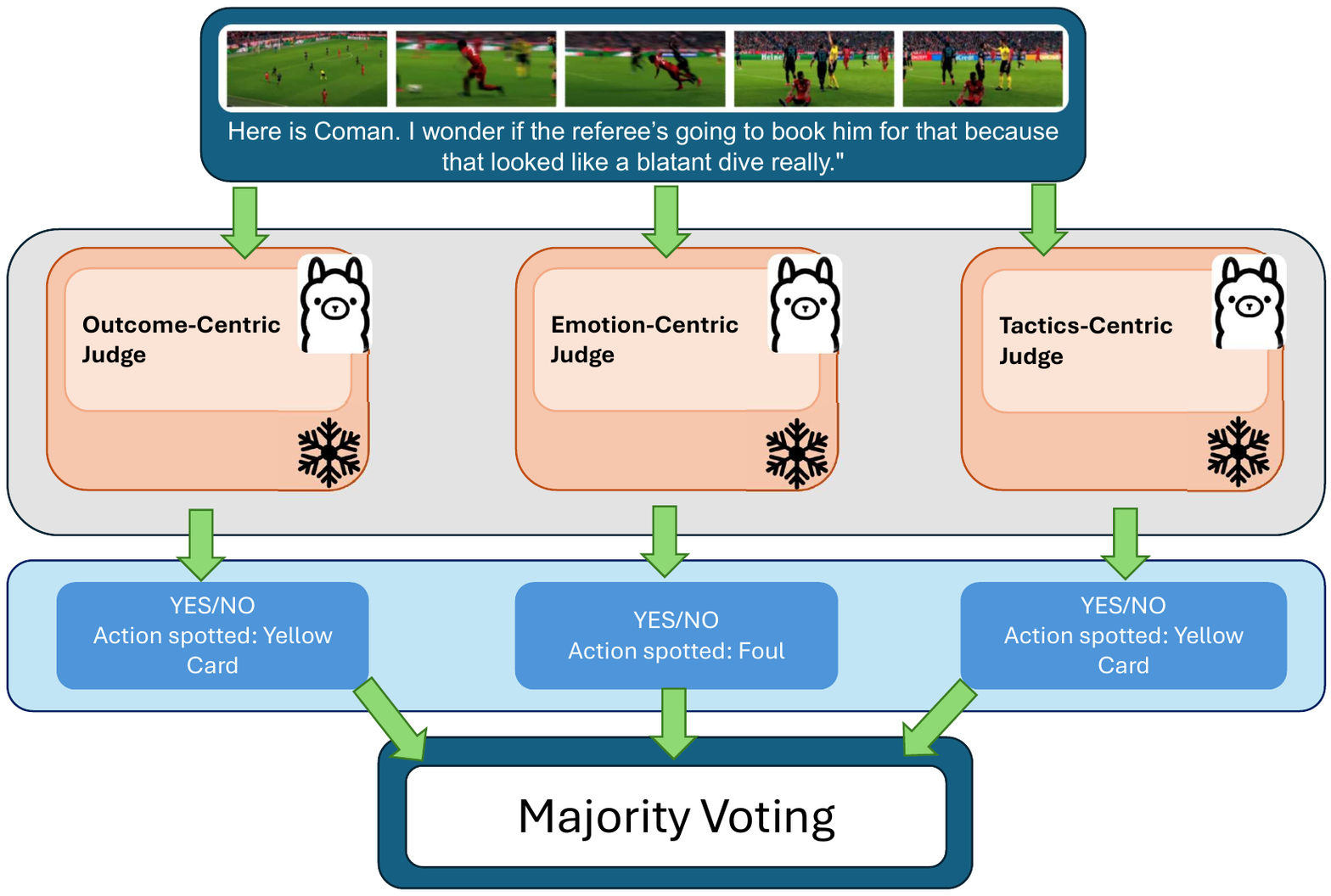}
    \caption{Our proposed LLM-based action spotting framework.}
    \label{fig: pipeline}
\end{figure}
% We define two key research questions (RQs) for our study:

% \begin{tcolorbox}[colframe=blue!60, colback=blue!5!white, coltitle=black, title=Research Question 1]
% \textbf{Can fine-grained commentary provide sufficient information for accurate action spotting?}  
% In this study, we investigate whether the detailed descriptions provided by commentators are enough to reliably detect key actions, such as goals, cards, and substitutions, without relying on visual data.
% \end{tcolorbox}

% \begin{tcolorbox}[colframe=green!60, colback=green!5!white, coltitle=black, title=Research Question 2]
% \textbf{How can we leverage Large Language Models (LLMs) for event detection in textual commentary?}  
% We examine how a system of three LLMs, each specialized in outcome, excitement, and tactics, can be used to evaluate timestamped commentary windows and identify event-worthy segments.
% \end{tcolorbox}
\begin{itemize}
    \item We redesign action spotting as a text based task as compared to a visual based task, utilising the Soccernet-Echoes dataset \cite{SN-ECHOES}.
    \item We design and implement a three-LLM system that judges the commentary based on outcome , excitement, and tactics.
    \item We demonstrate that expert commentary, in many cases, provides comparable information for event detection compared to visual cues.
    \item We show that, by focusing on commentary alone, it is possible to detect key events reliably, highlighting the potential of language-centric models for sports analytics.
\end{itemize}

The rest of the paper is structured as follows. Section \ref{sec: related-works} discusses present literature around using text-based inputs for video tasks and action spotting in soccer matches. Section \ref{sec: methodology} explains our proposed framework in detail. Section \ref{sec: eval} sheds light on the experimental setup and quantitative results. Finally we discuss some limitations in Section \ref{sec: limitations}.

\section{Related Work}
\label{sec: related-works}
\paragraph{Detailed Descriptions in Video-Based Tasks.}
In video-based understanding tasks, traditional models have primarily relied on visual features extracted from video frames to detect and classify events. However, recent research has begun exploring the use of fine-grained descriptions, specifically, textual information derived from transcriptions or commentary, to enhance performance in tasks like action spotting and event detection. \citet{xie2019visual} demonstrated that integrating visual information with text can improve performance in action recognition tasks, as descriptive cues often convey context that is missed by raw visual data.In addition, \citet{su2012crowdsourcing} highlighted the utility of crowd-sourced commentary to aid in object detection tasks, which suggests that action spotting in dynamic environments, such as sports, could be enhanced by considering detailed narrative descriptions. 
Recent work shows that textual descriptions can carry action semantics the pixels miss and when transcribed reliably, can act as a compact surrogate for frames. For soccer specifically, dense, timestamped commentary corpora like SoccerNet-Caption \cite{Mkhallati_2023} and GOAL \cite{qi2023goal} establish the feasibility of commentary-anchored modeling, while MatchTime \cite{rao-etal-2024-matchtime} highlights and fixes video-text misalignment—a key pain point for using commentary in downstream tasks. Robust ASR models such as Whisper \cite{radford2022robustspeechrecognitionlargescale} makes multi-accent, broadcast-noise transcripts viable at scale, strengthening the case for text-first pipelines. 

% [Added] Beyond soccer, older and adjacent lines of work have leveraged live text/commentary for event detection and summarization, suggesting long-standing signal value in narration (Halin et al., 2012
% ; Zhang et al., ACL’16
% ).
% These works point toward the potential of substituting or supplementing visual cues with textual descriptions, thereby providing more precise context for video interpretation tasks.

\paragraph{Action Spotting in Soccer Videos.}
Action spotting in soccer has long relied on visual inputs, particularly tracking players and ball movements. However, recent developments in leveraging commentary and other textual sources for action detection have gained attention. \citet{giannakopoulos2016soccer} proposed a method that uses timestamped commentary as input to detect key moments in soccer, such as goals or penalties, demonstrating that textual data can complement traditional visual cues. Another approach by \citet{andrews2024aicommentator} used a multi-modal network that combines both video frames and textual commentary to detect key events in football matches.
The SoccerNet benchmark \cite{deliege2021soccernet} formalized spotting as timestamp localization, driving a largely video-first literature . Classical baselines learn visual features and pool them temporally such as CALF \cite{cioppa2020contextawarelossfunctionaction} and NetVLAD++ temporal pooling \cite{giancola2021temporally} . Subsequent models improved localization via stronger heads/sequence learning, including RMS-Net \cite{RMSNET} and compact E2E-Spot \cite{hong2022spotting}. Recent transformer systems such as ASTRA \cite{ASTRA} push tight-tolerance accuracy further and even add audio for non-visible cues. Broader universal efforts such as UniSoccer \cite{rao2025towards} argue for richer taxonomies and multi-task foundations that still place video at the center. These threads collectively set a strong video baseline for action spotting.
\\

Despite these promising advancements, there remains a gap in fully utilizing fine-grained commentary for video understanding tasks like action spotting, especially in the context of soccer. Existing methods either rely on computationally expensive visual cues or fail to achieve consistent performance with textual input alone. 
% In this paper, we address this gap by proposing a novel method that employs a three-LLM (Large Language Model) judge setup for action spotting based purely on textual commentary.

\section{Methodology}
\label{sec: methodology}

\paragraph{Large Language Model Judges.}
We use Llama 3.1 8B \cite{grattafiori2024llama3herdmodels} to instantiate three specialized judges that operate over a shared label space of  the 17 SoccerNet-V2 classes and \textsc{NO-ACTION}. Each judge sees the same 10\,s commentary window (5\,s stride) but is prompted with a distinct \emph{evidence lens} (Outcome, Tactics, Emotion). All three judges return a single class (or \textsc{NO-ACTION}) and confidence score. Judges are steered by a dedicated system prompt and 2--3 few-shot exemplars .

\begin{tcolorbox}[
  width=\columnwidth, colback=gray!3, colframe=black!60, boxrule=0.45pt,
  arc=1.5pt, left=4pt, right=4pt, top=3pt, bottom=3pt]
\centering
\setlength{\parskip}{3pt}

{\large\faTrophy\enspace\textbf{Outcome-centric Judge}}\\[-2pt]
\small Prioritizes refereeable outcomes (goal, penalty, yellow/red), explicit referee phrases.

{\large\faChessKnight\enspace\textbf{Tactics-centric Judge}}\\[-2pt]
\small Emphasizes set-pieces and structure (corner, free-kick, substitution, formation/press).

{\large\faBolt\enspace\textbf{Emotion-centric Judge}}\\[-2pt]
\small Uses rhetorical intensity and urgency to resolve ambiguous cases; conservative when negations appear (``over the bar'', ``flag is up'').
\vspace{2pt}

\end{tcolorbox}

\begin{tcolorbox}[
  colback=blue!2, colframe=blue!50!black, boxrule=0.45pt, arc=1.6pt,
  left=3pt, right=3pt, top=2pt, bottom=2pt]
\small
\textbf{Input:} full English commentary for a 10\,s window (5\,s stride).\\
\textbf{Output (per judge):}\\
\begin{enumerate}[nosep,leftmargin=*,label=\textbf{\arabic*.}]
  \item A \textbf{single label} in \{17 SoccerNet-V2 classes\} $\cup$ \{\textsc{NO-ACTION}\}.
  \item A \textbf{confidence} in $[0,1]$ (calibrated from model's self-score).\\
\end{enumerate}
\texttt{{Abstention is expressed as \textsc{NO-ACTION}; we use higher thresholds for the Emotion judge to avoid rhetorical over-triggering.}}
\end{tcolorbox}

% \begin{tcolorbox}[colframe=black, colback=yellow!10!white, title=Why look beyond Outcome?] For many soccer matches, emotional moments such as goals and fouls are often accompanied by an excited tone in the commentary. This emotional context is a strong signal for the excitement judge, which can detect such moments without needing to rely on visual cues. The strategic context provided by the tactical judge is similarly informed by phrases that suggest changes in team strategy or formation. \end{tcolorbox}
% \begin{figure}
%     \centering
%     \includegraphics[width=1\linewidth]{soccernet_actions_plot.png}
%     \caption{A detailed look into the 17 action classes for Soccernet-v2.}
%     \label{fig:soccernet-actions}
% \end{figure}
\paragraph{Majority Voting System.}
Once each judge makes its decision, we aggregate the results using a majority voting mechanism. If at least two of the three judges agree on the presence of a relevant action, the action is considered "spottable" and is classified as an event worthy of attention. If the judges disagree, the action is not classified as relevant. This ensures that only the most unanimously recognized actions are selected.

\paragraph{Out-of-World Action Classification.}
In addition to the 17 predefined action classes, our system is designed to handle "out-of-world" actions—those that may be noteworthy but do not fall under any of the predefined classes. For instance, a player might execute a spectacular skill move or a controversial non-foul action, which can be exciting and relevant but doesn’t match the typical goal or penalty. In this case, the judges are given the opportunity to classify the action as "out of world," providing a broader view of game dynamics that goes beyond standard categories.

\begin{table}[t]
\centering
\tiny
\resizebox{\columnwidth}{!}{%
\begin{tabular}{@{}l l cc@{}}
\toprule
\textbf{Method} & \textbf{M} & \textbf{mAP (\%)} & \textbf{Tight mAP (\%)} \\
\midrule
CALF~\cite{cioppa2020contextawarelossfunctionaction} & Video & 49.7 & -- \\
RMS-Net~\cite{RMSNET} & Video & 63.49 & 28.83 \\
FCMA~\cite{zhou2021feature} & Video & 73.77 & 47.05 \\
E2E-Spot (RegNetY-200MF)~\cite{hong2022spotting} & Video & 73.25 & 61.19 \\
E2E-Spot (RegNetY-800MF)~\cite{hong2022spotting} & Video & 74.05 & 61.82 \\
ASTRA~\cite{ASTRA} & Video & \textbf{78.09} & \textbf{66.82} \\
\midrule
\textbf{Random Text-Only (ours, baseline)} & Text & \textbf{12.0} & \textbf{10.5} \\
\textbf{LLM-Based (Ours)} & \textbf{Text} & \textbf{64.5} & \textbf{60.8} \\
\bottomrule
\end{tabular}%
}
\caption{mAP and tight mAP on SoccerNet-v2 for video- vs text-based pipelines. M = Modality (Video/Text).}
\label{table:results}
\end{table}
% Requires: \usepackage{booktabs} and \usepackage{graphicx}
\begin{table*}[h!]
\centering
\resizebox{\textwidth}{!}{%
\begin{tabular}{@{}l l r r r r@{}}
\toprule
\textbf{Method} & \textbf{Input} & \textbf{FLOPs/frame (GF)} & \textbf{Total FLOPs (TF)} & \textbf{Time/frame (ms)} & \textbf{Time/match (sec)} \\
\midrule
RegNetY-200MF (E2E-Spot)    & Video & 0.20  & 2.16   & 0.3   & 3.24 \\
RegNetY-800MF (E2E-Spot)    & Video & 0.80  & 8.64   & 0.9   & 9.72 \\
ResNet-152 (baseline feats) & Video & 11.5  & 124.2  & 1.8   & 19.44 \\
R(2+1)D (3D CNN)            & Video & --    & --     & 11.0  & 118.8 \\
\midrule
\textbf{Ours: LLM (text-only)} & \textbf{Text} & \textbf{--} & \textbf{--} & \textbf{--} & 146.5 \\
\bottomrule
\end{tabular}%
}
\caption{Efficiency comparison on FLOPs and wall-clock GPU/CPU time needed for a full match evaluation.}
\label{tab:efficiency}
\end{table*}

\section{Evaluation and Results}
\label{sec: eval}

\paragraph{Setup.}
We evaluate on the SoccerNet-v2 test split using the SoccerNet-Echoes commentary \cite{SN-ECHOES} as input. Events follow the 17-class SoccerNet taxonomy. Our system operates on 10\,s windows (5\,s stride) and uses three Llama\,3.1\,8B judges. We report the official average-mAP (``mAP'') over $\delta\!\in\![5,60]$\,s and tight Average-mAP (``Tight'') over $\delta\!\in\![1,5]$\,s in \ref{table:results}. For efficiency (Table~\ref{tab:efficiency}) we normalize video compute to a \textbf{90\,min match at 2\,FPS} (10{,}800 frames) and report backbone FLOPs/frame and measured per-frame time on an A5000 from prior work \cite{hong2022spotting} and for our text-only system we report CPU wall-clock time per match.

\paragraph{Textual Random Baseline.}  
To establish a strict lower bound we create a commentary-anchored randomness baseline that predicts actions without reading the text.  For each commentary sentence $s_k=(\tau_k,\ell_k,\text{text}_k)$ in a half we sample a Bernoulli coin for every action class $c$ with probability equal to that class’s empirical commentary prior $p_c$ (estimated on the train split).  If the coin succeeds we emit a pseudo detection $(\tau_k,c,0.5)$; overlapping detections of the same class within $2\delta$ ($\delta{=}10$ s) are merged by keeping the earliest.  This design respects the real timestamp distribution yet ignores all lexical information, yielding the hardest chance-level floor against which any text-aware model must improve.

\paragraph{Main results.}
Table~\ref{table:results} shows that our text-only system achieves \textbf{64.5 mAP} and \textbf{60.8 Tight}, substantially outperforming the \textit{Random Text-Only} baseline (12.0 / 10.5) and approaching recent video methods despite using no visual frames. Relative to strong video pipelines, we are close on the tight metric (\textbf{60.8} vs \textbf{61.82} for E2E-Spot RegNetY-800MF; \textbf{60.8} vs \textbf{66.82} for ASTRA), while trailing more on loose mAP (\textbf{64.5} vs \textbf{74.05} and \textbf{78.09}). Compared to RMS-Net, our tight score is more than \textbf{2$\times$} higher (60.8 vs 28.83) and our loose mAP is competitive (64.5 vs 63.49). The pattern aligns with the nature of commentary: explicitly lexicalized, refereeable outcomes (goals, penalties, bookings, substitutions) are well localized in time, benefitting Tight mAP; at larger tolerances we remain intentionally conservative via abstention, trading some recall for precision.

\paragraph{Efficiency ablation.}
Table~\ref{tab:efficiency} compares per–90\,min match compute and explains our savings. Video pipelines pay a cost that scales with the number of \emph{visual tokens}; text scales with \emph{text tokens}. Let $F$ be frames per match and $P$ the patch tokens per frame (ViT-style). Then
$\text{visual-tokens} = F\!\times\!P$ and $\text{text-tokens} = N_t$.
At 2\,FPS, $F{=}10{,}800$. For ViT-B/16 at $224^2$, $P{=}(224/16)^2{=}196$, so $F\!\times\!P\!\approx\!2.12\times 10^6$ visual tokens/match, whereas ASR produces only $N_t=\mathcal{O}(10^4)$ text tokens—two orders of magnitude fewer. Even with CNNs (no explicit patches), the effective per-frame compute (GFLOPs/frame) still scales with $F$ and dominates.

At 2\,FPS, published video backbones span 2.16–124.2\,TFLOPs per match and 0.3–1.8\,ms per frame on an A5000 (3.24–19.44\,s per match; a 3D CNN is 118.8\,s). Our pipeline performs no video feature extraction (zero video FLOPs) and instead scales with $N_t$ and LLM tokens/s. On CPU, our measured end-to-end time for a full match is 146.5\,s (2.44\,min), removing the dominant frame-processing term and any GPU requirement.

% \noindent\textit{Why this is better in practice.}
% \begin{itemize}[leftmargin=1.2em,itemsep=2pt,topsep=2pt]
%   \item \textbf{Zero-frame budget.} Removing $F$ from the pipeline (no decoding, no storage, no frame features) collapses the dominant compute and IO term; operationally simpler and cheaper at scale.
%   \item \textbf{CPU-deployable.} A full match in \textbf{146.5\,s} on CPU makes batch processing practical without accelerators; trivially parallelizable across cores/matches.
%   \item \textbf{Tight accuracy retained.} Despite zero frames, we match or approach strong video systems on the \emph{tight} metric (Table~\ref{table:results}), which is the localization-sensitive regime.
%   \item \textbf{Predictable scaling.} Complexity grows with $N_t$ (commentary length), not with pixels or patches; this is stable across broadcasts and leagues.
% \end{itemize}

% \noindent\textit{Fairness note.} Video times in Table~\ref{tab:efficiency} are \emph{GPU backbone-only} (A5000), excluding decoding/IO, multi-backbone ensembles, and post-processing; our number is \emph{CPU end-to-end}. The two columns answer different questions: best-case GPU backbone speed vs.\ actual CPU time to process a match. We also report FLOPs/match for video to expose the large frame-processing budget that our text pipeline removes outright.

\paragraph{Discussion.}
(1) \textbf{Tight localization from text.} When outcomes are spoken (“penalty given”, “booked”, “and it’s in”), the language signal is temporally sharp, explaining our proximity to video SOTA on Tight mAP.  
(2) \textbf{Loose-gap sources.} Non-verbal micro-events and terse restarts are under-described in commentary, which hurts loose recall and favors video.  
(3) \textbf{Design effects.} Confidence thresholds and majority voting suppress rhetorical false positives (near-misses), improving precision; temporal NMS converts overlapping window votes into a single timestamp per event.  
(4) \textbf{Compute and deployment.} Zero-frame processing plus competitive Tight mAP make the approach attractive for CPU-scale batch processing (clubs/broadcasters) and for low-cost inference at volume.

\section{Limitations}
\label{sec: limitations}
While our framework shows promising results, there are several limitations to consider. First, the performance of our system is heavily dependent on the quality of the commentary and transcription. Inaccurate or incomplete commentary can hinder the ability of our judges to correctly identify action-worthy events, leading to lower accuracy in the action spotting task. Similarly, the quality of transcription performed by Whisper plays a critical role. Errors in the transcription process can result in incorrect words or misplaced timestamps, directly affecting the action spotting metrics, including mean Average Precision (mAP). These transcription errors could affect the reliability of the timestamped actions and ultimately influence the results of the semantic judging.
Additionally, our framework assumes that the provided commentary is sufficiently detailed and relevant for the action spotting task. In cases where the commentary lacks context or important details, the system's performance may degrade. We aim to address this in our future work. 

%The scalability of the system could be a concern when applying the framework to longer or more complex matches with a high volume of commentary.


\bibliography{acl_latex}

% \appendix

% \section{Appendix}
% \label{sec:appendix}

%This is an appendix.

\end{document}